%% file: main.tex
\documentclass[lettersize,journal]{IEEEtran}
\usepackage{graphicx} % Required for inserting images
\usepackage{multirow}
\usepackage{amsmath}
\usepackage[dvipsnames]{xcolor}
\usepackage{url}
\usepackage{pifont}
\usepackage{pbox}
\usepackage{subcaption}

\newcommand*{\TR}[1]{\textcolor{black}{#1}}
\newcommand*{\RR}[1]{\textcolor{black}{#1}}

\newcommand{\cmark}{{\ding{51}}}
\newcommand{\xmark}{{\ding{55}}}

\begin{document}
\title{Representation Learning with Parameterised Quantum Circuits for Advancing Speech Emotion Recognition}
\author{Thejan Rajapakshe$^{*}$,~\IEEEmembership{Student Member,~IEEE}, Rajib Rana,~\IEEEmembership{Member,~IEEE}, Farina Riaz, \\
Sara Khalifa,~\IEEEmembership{Member,~IEEE}, and Bj\"{o}rn W.\ Schuller,~\IEEEmembership{Fellow,~IEEE}
\thanks{Thejan Rajapakshe \& Rajib Rana are with the University of Southern Queensland, Australia}
\thanks{Farina Riaz is with Commonwealth Scientific and Industrial Research Organisation, Australia}
\thanks{Sara Khalifa is with Queensland University of Technology, Australia}
\thanks{Bj\"{o}rn W.\ Schuller is with CHI -- Chair of Health Informatics, Technical University of Munich, Germany, and GLAM -- the Group on Language, Audio, \& Music, Imperial College London, UK}
\thanks{$^{*}$Thejan.Rajapakshe@unisq.edu.au}% <-this % stops a space
\thanks{Manuscript received January XX, 2025; revised January XX, 2025.}}

\maketitle
\IEEEpeerreviewmaketitle

\begin{abstract}

\RR{Quantum machine learning (QML) offers a promising avenue for advancing representation learning in complex signal domains. In this study, we investigate the use of parameterised quantum circuits (PQCs) for speech emotion recognition (SER)—a challenging task due to the subtle temporal variations and overlapping affective states in vocal signals. We propose a hybrid quantum-classical architecture that integrates PQCs into a conventional convolutional neural network (CNN), leveraging quantum properties such as superposition and entanglement to enrich emotional feature representations. Experimental evaluations on three benchmark datasets—IEMOCAP, RECOLA, and MSP-IMPROV—demonstrate that our hybrid model achieves improved classification performance relative to a purely classical CNN baseline, with over 50\% reduction in trainable parameters. This work provides early evidence of the potential for QML to enhance emotion recognition and lays the foundation for future quantum-enabled affective computing systems.}

\end{abstract}


\input{10_introduction}

\input{20_related_work}

% \input{25_background}

\input{30_methodology}

\input{35_experimental_setup}

\input{40_results}

\input{50_discussion}

\input{60_conclusion}

\bibliographystyle{IEEEtran}
\bibliography{references}

\end{document}

%% file: 10_introduction.tex
\section{Introduction}

% \IEEEPARstart{R}{}epresentation learning has emerged as a dominant approach in machine learning, enabling systems to extract meaningful features directly from raw data. This paradigm is particularly significant in the field of Speech Emotion Recognition (SER) as it addresses the complex task of interpreting emotional expressions conveyed through speech. Despite recent advancements, the field still faces challenges in achieving reliable accuracy without complicating the model with high number of parameters. Key factors contributing to these difficulties include the overlapping nature of emotional states, variability among speakers, and the intricate dependencies among various speech features, as noted in the literature~\cite{ShahFahad2021AEnvironment,Daneshfar2021SpeechImbalance,George2024ANoise}.

\IEEEPARstart{R}{}epresentation learning has become a foundational approach in machine learning, enabling systems to extract meaningful patterns directly from raw data. In the field of Speech Emotion Recognition (SER), this paradigm is particularly valuable for interpreting the rich and often ambiguous emotional signals embedded in vocal expressions. Despite recent progress, SER continues to face significant challenges in achieving high accuracy without resorting to increasingly complex models with a large number of trainable parameters. Key obstacles include the overlapping nature of emotional states, speaker variability, and the intricate interdependencies among acoustic features such as pitch, energy, and rhythm~\cite{ShahFahad2021AEnvironment,Daneshfar2021SpeechImbalance,George2024ANoise}.

\RR{Quantum Machine Learning (QML) offers a compelling alternative by exploiting the unique computational characteristics of quantum systems: most notably, superposition and entanglement. Superposition enables quantum models to process multiple states simultaneously, allowing for a more holistic analysis of the nuanced and concurrent variations inherent in emotional speech. Rather than evaluating features in isolation or sequentially, as is typical in classical systems, quantum models can assess complex patterns across all dimensions in parallel.}

\RR{Entanglement further enhances this potential by capturing the dependencies between speech features that classical models often overlook or treat independently~\cite{Sharma2023TheClassification,Wang2024TransitionLearning,WANG2024ExploringLearning,Gaspar2024Entanglement-enhancedSingle-Photons}. In practice, emotional states are rarely expressed through isolated vocal characteristics; instead, they emerge from subtle relationships between multiple attributes, such as correlations between tone, tempo, and energy. QML systems inherently encode and analyse these relationships, leading to richer and more robust representations of affective information.}

\RR{Parameterised Quantum Circuits (PQCs) are among the most promising constructs in QML. These circuits consist of tunable quantum gates, which can be trained to optimise complex representations. Unlike conventional deep neural networks—which often rely on large numbers of parameters to model high-dimensional spaces—PQCs can encode exponentially large feature spaces using comparatively fewer resources~\cite{Farhi2018ClassificationProcessors}. This makes them especially suitable for SER, where modelling emotional intricacies efficiently is critical.}

\RR{The emergence of Noisy Intermediate-Scale Quantum (NISQ) devices, such as quantum processors with limited qubit counts and moderate noise resilience, has made it feasible to explore hybrid quantum-classical learning strategies. While current quantum hardware is not yet mature for large-scale deployment, simulations of PQC-based models offer valuable insights into their performance and architectural viability under NISQ constraints.}

\begin{figure}[ht]
    \centering
    \includegraphics[width=1\linewidth]{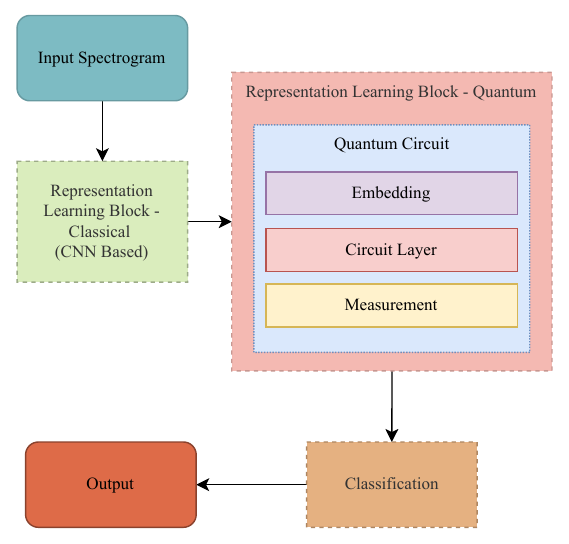}
    \caption{A simplified architecture of the proposed Hybrid Classical-Quantum SER model.}
    \label{fig:model_architecture_small}
\end{figure}

\RR{In this study, we propose a novel hybrid architecture for speech emotion recognition that integrates PQCs within a classical convolutional neural network framework, as illustrated in Figure~\ref{fig:model_architecture_small}. By combining the expressive power of quantum circuits with the proven feature extraction capabilities of CNNs, our approach addresses key limitations of classical models in terms of both accuracy and computational efficiency. Experiments conducted on three benchmark datasets—IEMOCAP, RECOLA, and MSP-IMPROV—demonstrate that the hybrid model achieves improved classification performance compared to a classical CNN baseline, while reducing the number of trainable parameters by over 50\%.}

\RR{This work contributes to the growing field of quantum-enhanced affective computing by demonstrating that PQCs can be effectively applied to emotionally rich, temporally structured data. It also provides empirical insights into how quantum circuit design, particularly expressibility and entanglement, affects model performance, laying the groundwork for future implementation on real quantum hardware.}

% The key contributions of this study are as follows:

% \begin{itemize}
%     \item A novel SER hybrid classical-quantum framework that integrates PQCs with conventional CNN. This architecture leverages quantum superposition and entanglement to enhance feature representation, addressing the intricate dependencies and overlapping emotional states inherent in speech signals.
%     \item Compared to purely classical approaches, the proposed model achieves superior performance in both binary and multi-class emotion classification tasks while significantly reducing the number of trainable parameters of the model.
%     \item Experimental validation across multiple benchmark datasets (IEMOCAP, RECOLA, and MSP-Improv) underscores the efficacy of the hybrid model. The results indicate improved accuracy with smaller number of trainable parameters, setting a strong foundation for future exploration of quantum-enhanced SER applications.
% \end{itemize}

%% file: 20_related_work.tex
\section{Related Work}

Speech Emotion Recognition (SER) has experienced significant advancements through the adoption of representation learning, a paradigm where models automatically extract meaningful features directly from raw speech data. Deep learning has been instrumental in this shift, enabling the development of models capable of learning rich and task-specific representations, surpassing the limitations of traditional hand-engineered features. However, despite improving accuracy and adaptability, deep learning-based methods face challenges, including high computational demands and difficulties in capturing intricate relationships within high-dimensional feature spaces. Quantum machine learning (QML) offers a promising solution by leveraging quantum circuits to efficiently process and represent complex, high-dimensional data with fewer resources. This section explores recent advancements in representation learning for SER and examines how QML can address the computational and modelling challenges of classical deep learning approaches.

\subsection{Representation Learning in SER}
 Representation learning has emerged as a powerful paradigm for extracting meaningful features from speech data, enabling significant advancements in SER systems.

\subsubsection{Automatic Feature Extraction and Task Specificity} Conventional feature engineering requires prior domain knowledge to design features, which may not generalise well across datasets or tasks. Representation learning, in contrast, automates feature extraction by learning task-relevant features directly from raw audio signals. Trigeorgis et al. demonstrated that deep convolutional recurrent networks could extract hierarchical features from raw waveforms, outperforming traditional MFCC-based systems~\cite{Trigeorgis16-AFE}. This approach eliminates the need for manual feature design, reducing human intervention and improving adaptability to diverse datasets~\cite{Stolar2018RealLearning, Pan2024GFRN-SEA:Analysis, Zhang2024MultimodalSinc-convolution}.

\subsubsection{Enhanced Representational Power}
 Deep learning architectures like Convolutional Neural Networks (CNNs) and Long Short-Term Memory (LSTM) networks learn temporal and spectral dependencies, enhancing their ability to model subtle emotional cues~\cite{Etienne2018CNN+LSTMAugmentation, RayhanAhmed2023AnRecognition}. Satt et al.~\cite{Satt2017EfficientSpectrograms} found that deep models with spectrogram-based inputs significantly outperformed systems using hand-crafted features to detect speech emotions. 
 %BS: editing for language directly...
 Zhao et al.~\cite{Zhao2019SpeechNetworks} introduced two CNN-LSTM architectures: one utilising 1D CNNs for raw audio input and the other employing 2D CNNs for Mel-spectrogram inputs. They reported achieving an accuracy of 52.14\% on the IEMOCAP database for speaker-independent SER using the 2D CNN-LSTM model.

 \TR{Further advancing classical methods, researchers have also explored complex architectures to address multiple SER challenges simultaneously. For instance, Daneshfar and Kabudian proposed a system using a hierarchical multi-layer sparse auto-encoder combined with an Extreme Learning Machine (ELM)~\cite{Daneshfar2021SpeechImbalance}. Their approach utilised a rich set of spectral and spectro-temporal features and notably introduced a new adaptive weighting method specifically to counteract the problem of data imbalance, a persistent challenge in SER datasets. This highlights a direction in classical SER research focused on creating sophisticated, multi-stage processing pipelines to extract discriminative features and build robust classifiers.}

\subsubsection{Scalability Across Languages and Domains}
Hand-crafted features often need to be tailored for specific languages or domains, limiting their scalability. Representation learning, however, leverages large-scale pre-training to learn universal speech representations applicable across multiple tasks and languages. Huang et al.~\cite{Xie2019SpeechLSTM} highlighted that attention-based models trained on multilingual datasets generalise well, enabling SER systems to scale efficiently.
Mirsamadi et al.~\cite{Mirsamadi2017AutomaticAttention} demonstrated that deep recurrent neural networks can effectively learn and aggregate frame-level acoustic features into compact utterance-level representations for improved emotion recognition. Additionally, they proposed a local attention-based pooling strategy to focus on emotionally salient regions of speech, achieving superior performance on the IEMOCAP corpus compared to existing methods.

% Representation learning has brought transformative advancements to SER, enabling the automatic extraction of task-relevant features and outperforming traditional hand-crafted approaches~\cite{Li2019TowardsRecognition, Latif2023SurveyRecognition}. However, the underlying deep learning models are computationally intensive, leading to significant resource demands, especially during training. Furthermore, classical deep learning struggles to efficiently model certain complex correlations and entanglements in high-dimensional feature spaces, which are critical for capturing subtle emotional cues in speech~\cite{Jahangir2021DeepChallenges}.

\subsubsection{Robustness to Limited Labelled Data}
Representation learning, especially self-supervised approaches, reduces the reliance on labelled data, which is often scarce in SER. Models pre-trained on large unlabelled datasets, such as HuBERT~\cite{Hsu2021HuBERT:Units}, can be fine-tuned with limited labelled data, making them ideal for settings with sparse annotations. This adaptability is a significant advantage over traditional methods requiring extensive labelled data for optimal performance.

\subsubsection{Challenges in Representation Learning for SER: Computational Demands and High-Dimensional Modelling}
Representation learning has brought transformative advancements to SER, enabling the automatic extraction of task-relevant features and outperforming traditional hand-crafted approaches~\cite{Li2019TowardsRecognition, Latif2023SurveyRecognition}. However, the underlying deep learning models are computationally intensive, leading to significant resource demands, especially during training. Moreover, the classical deep learning models face challenges in efficiently modelling complex correlations and entanglements in high-dimensional feature spaces, which are critical for capturing subtle emotional cues in speech~\cite{Jahangir2021DeepChallenges}.

\subsubsection{QML addressing the Challenges in Representation Learning for SER  }
QML offers a promising alternative to address these issues. Quantum circuits, by leveraging principles of superposition and entanglement, have the potential to represent and process exponentially large feature spaces with fewer resources compared to classical systems~\cite{Gong2024QuantumCircuits}. This capability could enable more efficient representation learning for SER, particularly in scenarios with limited labelled data or high-dimensional inputs.

\subsection{Quantum Machine Learning for SER and Related Fields}
Early QML research focused on foundational algorithms like the quantum support vector machine~\cite{Rebentrost2014QuantumClassification, Biamonte2017QuantumLearning} and the Harrow-Hassidim-Lloyd (HHL) algorithm for solving linear systems~\cite{Biamonte2017QuantumLearning}. With the advent of noisy intermediate-scale quantum (NISQ) devices, practical quantum-enhanced models, such as PQCs, have become feasible~\cite{Cerezo2021VariationalAlgorithms}. These advancements suggest that QML can potentially overcome the limitations of classical deep learning, particularly in modelling high-dimensional data and reducing resource requirements.

\subsubsection{Parameterised Quantum Circuits (PQCs)}
QML has also shown promise in augmenting traditional neural network structures through Quantum Neural Networks (QNNs). By employing PQCs as neural layers, QNNs can encode quantum states while exploiting quantum advantages for specific tasks. Farhi and Neven pioneered quantum-enhanced learning frameworks for combinatorial optimisation~\cite{Farhi2018ClassificationProcessors}, and Schuld and Killoran demonstrated how quantum circuits can efficiently process and generalise high-dimensional data~\cite{Schuld2019QuantumSpaces}, showcasing their suitability for complex learning tasks. 
% While none of the above studies focused on SER, Qu et al.\ introduced a Quantum Federated Learning (QFL) algorithm for emotion recognition in speech, designed with applications in Internet-on-Vehicles (IoV) technology. Their method employed quantum minimal gated unit RNNs, emphasising privacy and robustness in noisy environments~\cite{Qu2024QFSM:IoV}. While this work used PQCs, it did not focus on representation learning.
\TR{While none of the above studies focused on SER, Qu et al. introduced a Quantum Federated Learning (QFL) algorithm for emotion recognition in speech. Their method emphasised privacy and robustness in noisy environments by employing quantum minimal gated unit RNNs~\cite{Qu2024QFSM:IoV}. However, a key shortcoming of this work is that it did not focus on representation learning. By not leveraging the feature representation capabilities of PQCs, the model's ability to capture the complex, high-dimensional dependencies inherent in emotional speech data remained unexplored, limiting its potential effectiveness for the core SER task.}

\subsubsection{Hybrid Classical-Quantum Architectures}
Hybrid quantum-classical architectures have been explored to enhance model expressiveness and training efficiency. These architectures integrate the representational strengths of classical deep learning with the computational power of quantum systems~\cite{Liu2019HybridNetworks,Xiang2024QuantumDiagnosis}. Hybrid models are particularly suited for tasks where classical models face limitations, as quantum layers can introduce novel features and expand the solution space. 
% Thejha 
%BS: This looks weird - why Theja B. and not only a full last name? Please also check above "Zhiguo Qu" - usually, you only write last name et al. - please unify. Also for Michael Esposito and so on ---> Esposito et al. (!) - In the end, I did unify already :)
% et al.\ applied a hybrid quantum-classical CNN for speech recognition, demonstrating the potential of hybrid models in audio processing~\cite{Thejha2023SpeechNetwork}. However, their work did not specifically target SER tasks. Similarly, Esposito et al.\ employed QNNs to process MFCC features for audio classification, passing the learnt representations through an RNN for COVID-19 cough detection~\cite{Esposito2022QuantumHealthcare}. While this study utilised hybrid architectures and PQCs, it did not focus on representation learning for SER. 
\TR{Other explorations into hybrid quantum-classical architectures have shown promise but have not yet unlocked a quantum advantage for SER. For instance, Thejha et al. applied a hybrid quantum-classical CNN to general speech recognition, achieving performance merely comparable to classical CNNs without demonstrating a clear benefit~\cite{Thejha2023SpeechNetwork}. Likewise, work by Esposito et al. on audio classification for cough detection~\cite{Esposito2022QuantumHealthcare} also failed to investigate representation learning for SER. These studies, while foundational, stopped short of demonstrating that QML could either outperform classical methods or effectively model the nuanced feature representations required for robust emotion recognition.}
% Similarly, Norval and Wang proposed a Quantum AI-based approach for SER, utilising features like MFCC, pitch, and energy. Their system employed quantum circuits for encoding and variational quantum algorithms for classification, achieving an accuracy of 30\% on a custom dataset~\cite{Norval2024QuantumRecognition}. However, their approach did not investigate PQCs or representation learning for SER.
\TR{Similarly, Norval and Wang proposed a Quantum AI-based approach for SER, but its practical viability was limited, achieving a low accuracy of only 30\% on a custom dataset~\cite{Norval2024QuantumRecognition}. Their approach, which relied on basic quantum encoding and variational algorithms without exploring PQCs or representation learning, underscores the significant performance challenges that early QML-SER models faced.}

An area of significant interest is the application of hybrid architectures in CNNs, leading to the development of quantum CNNs (QCNNs). QCNNs effectively process structured data by utilising quantum circuits to perform convolution-like operations, as demonstrated by Cong et al.~\cite{Cong2019QuantumNetworks}.

\input{20_related_work_summary}

\subsection{Research Gap}
Table~\ref{tab:lit_summary} provides a comparative overview of existing research utilising QML for SER and related fields. While the table highlights investigations into hybrid classical-quantum architectures and PQCs within broader audio and speech processing domains, a significant research gap exists in their application to representation learning for SER. Three key observations characterise this gap:
\begin{itemize}
    \item Limited SER Focus - Existing QML research for audio predominantly targets broader applications like general audio classification and speech recognition, with comparatively little exploration of the specific nuances and complexities of SER.
    \item Unexplored Representation Learning - While some studies have incorporated hybrid classical-quantum models and PQCs in related audio tasks, their potential for representation learning within the SER domain remains largely untapped. This represents a significant avenue for future research.
    \item Potential for Advancement - This identified research gap highlights the novelty and substantial potential of integrating QML techniques, particularly PQCs and hybrid architectures, into representation learning paradigms for SER. Such integration promises to unlock more efficient and robust emotion recognition systems, addressing the current limitations of classical approaches.
\end{itemize}
This paper aims to investigate the of integration QML techniques, specifically PQCs and hybrid architectures, into representation learning frameworks to address these gaps and drive advancements in the field of  SER.
% [Add bullet points on the research contribution]

% Most prior research in QML for SER has concentrated on quantum-enhanced SVMs for classification, QFL, leaving a notable gap in exploring hybrid classical-quantum models for this purpose. To address this gap, our study combines the strengths of classical and quantum paradigms to assess their feasibility and effectiveness in enhancing SER performance. This work represents a pioneering effort in this area, validated through experiments on widely used SER datasets, including IEMOCAP, RECOLA, and MSP-Improv.

% Despite the growing interest in applying QML to various domains, our work explores a novel direction by utilising a hybrid classical-quantum model specifically for SER. While hybrid classical-quantum architectures have been employed in general audio and speech recognition tasks, their application to SER remains unexplored. 
% Previous efforts in QML for SER have primarily focused on using quantum-enhanced SVMs for classification tasks. This leaves a significant research gap in investigating the feasibility and effectiveness of hybrid classical-quantum models for SER. Our study addresses this gap by leveraging the strengths of both paradigms to evaluate their potential for improving SER performance, marking a pioneering effort in this domain. Also, this study validates the results by using popular SER datasets IEMOCAP, RECOLA, and MSP-Improv. 

%% file: 20_related_work_summary.tex
\begin{table*}[t]
\centering
\caption{Summary of the Literature related to Quantum (Q.) Machine Learning used for SER and related fields}
\label{tab:lit_summary}
\resizebox{\textwidth}{!}{%
\renewcommand{\arraystretch}{1.2}
\begin{tabular}{|p{2cm}|c|c|c|c|l|}
\hline
\multirow[c]{2}{*}{\textbf{Research Study}} & \multicolumn{4}{c|}{\textbf{Focus}} & \multicolumn{1}{c|}{\multirow{2}{*}{\textbf{Dataset}}}       \\ \cline{2-5}
{}  & \multicolumn{1}{p{2cm}|}{Hybrid Classical-Q. Model}  & \multicolumn{1}{p{1.8cm}|}{ Parameterised Q. Circuits } & \multicolumn{1}{p{2cm}|}{Representation Learning w/ Q.} & \multicolumn{1}{c|}{SER} & {}     \\ \hline
Esposito et al. 2022~\cite{Esposito2022QuantumHealthcare}           & \cmark        & \cmark  & \cmark      & \xmark  (Audio Classification)         & DiCOVA~\cite{Muguli2021DiCOVAAcoustics}                                                                                          \\ \hline
Thejha et al. 2023~\cite{Thejha2023SpeechNetwork}                         & \cmark        & \xmark  & \cmark      & \xmark (Speech Recognition)           & Not Available                                                                                             \\ \hline
Norval \& Wang 2024~\cite{Norval2024QuantumRecognition}     & \xmark        & \xmark  & \xmark (SVM)     & \cmark                                & Custom Dataset                                                                                        \\ \hline
Qu et al. 2024~\cite{Qu2024QFSM:IoV}                                 & \xmark        & \cmark  & \xmark (QFL)     & \cmark                                & \begin{tabular}[c]{@{}l@{}}CASIA~\cite{Tao2008DesignSpeech}\\ RAVDESS~\cite{Livingstone2018TheEnglish}\\ EMO-DB~\cite{Burkhardt2005ASpeech}\end{tabular}          \\ \hline
\textbf{This Paper}                                                          & \cmark        & \cmark  & \cmark      & \cmark                                & \begin{tabular}[c]{@{}l@{}}IEMOCAP~\cite{Busso2008IEMOCAP:Database} \\ RECOLA~\cite{Ringeval2013IntroducingInteractions} \\ MSP-Improv~\cite{Busso2017MSP-IMPROV:Perception}\end{tabular}                                                                               \\ \hline
\end{tabular}%{}
}
\renewcommand{\arraystretch}{1.0}
\end{table*}

%% file: 30_methodology.tex
\section{Contextualising Quantum Machine Learning for Speech Emotion Recognition}

In this study, we conduct a preliminary investigation to evaluate the applicability of Quantum Machine Learning (QML) with Parameterised Quantum Circuits (PQCs) for Speech Emotion Recognition (SER) tasks. By leveraging the quantum properties of superposition and entanglement, we hypothesise that QML can learn more complex and abstract representations from speech data, which are crucial for improving SER performance.

A simplified architecture of the proposed hybrid Classical-Quantum model is illustrated in Figure~\ref{fig:model_architecture_small}. In this model, the input spectrograms of speech audio data are first processed through a CNN-based representation learning block to generate classical representations. These representations are subsequently passed through a quantum representation learning block, where quantum operations are performed. Finally, the learnt quantum representations are fed into a classification block to predict the emotional state.

The Quantum Representation Learning block incorporates a quantum layer composed of three key components: the Quantum Embedding Component, the Quantum Circuit Layer, and the Quantum Measurement Component. This architecture allows the block to accept classical inputs, transform them into quantum states, and output classical representations, facilitating seamless representation transition in the hybrid classical-quantum model.

\subsection{Preamble}
\textbf{Quantum Embedding} refers to the process of encoding classical speech features into quantum states within the quantum Hilbert space. This step enables classical audio features to interface with quantum systems and serves as the foundation for quantum machine learning algorithms~\cite{Schuld2019QuantumSpaces}. To achieve quantum embeddings, PQCs or predefined encoding schemes are employed to transform speech feature inputs into quantum states that can be manipulated by quantum gates.

A \textbf{Quantum Circuit Layer} consists of a series of quantum gates that act on qubits to perform specific transformations or computations. Quantum circuits layers form the computational backbone of QNNs, enabling quantum systems to process information and learn. Two main types of circuits are utilised:\\
\textit{Static Quantum Circuits} -- These circuits have fixed configurations with pre-specified gate sequences and no adjustable parameters. Static circuits typically perform predefined operations, such as feature extraction or classical-to-quantum encoding, without further tuning during the training process. \\
\textit{Parameterised Quantum Circuits} -- PQCs include variable parameters -- such as rotation angles in quantum gates -- that are optimised iteratively during training~\cite{Cerezo2021VariationalAlgorithms}. As the trainable core of QNNs, PQCs enable dynamic adaptation to input data by minimising a defined cost function.

\begin{figure*}[!t]
\centering
\includegraphics[width=1\textwidth]{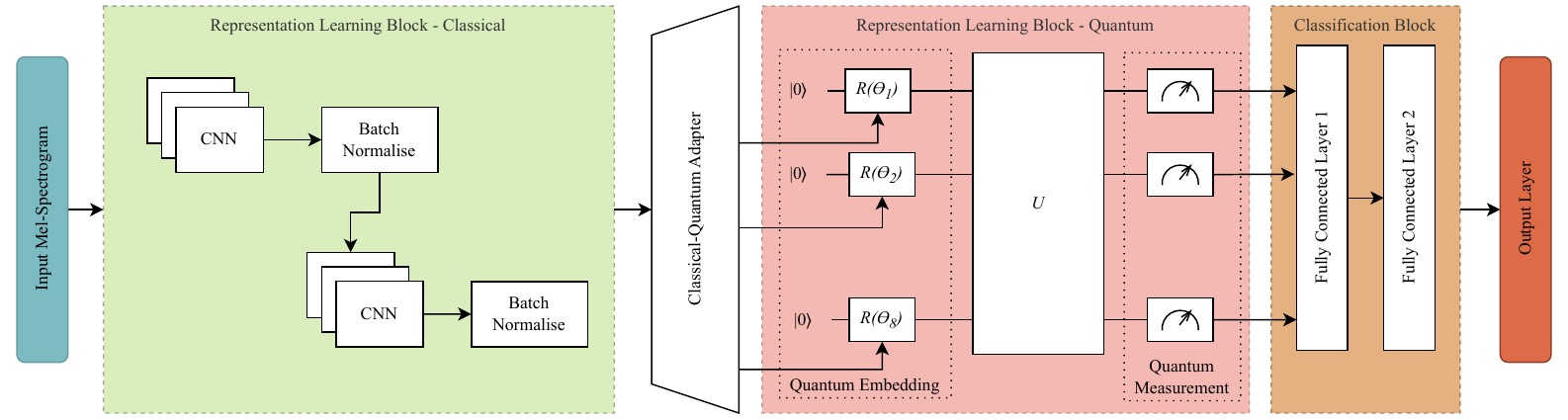}
\caption{Overview of the model architecture used in this study. Input features pass through a CNN representation learning module, followed by a Quantum representation learning module, before reaching the classification layer that assigns the input spectrogram to one of the output classes. The Quantum representation learning block contains Quantum Embedding, Quantum Circuit Layer (\textit{U}), and Quantum Measurement respectively.}
\label{fig:model_architecture}
\end{figure*}

\textbf{Quantum Measurement} is the process of extracting classical information from quantum states by projecting them onto specific basis states~\cite{Biamonte2017QuantumLearning}. In QNNs, measurements occur at the end of the quantum circuit to produce outputs such as classifications, probabilities, or projections. The choice of measurement basis significantly influences the interpretability and effectiveness of the results. In this study, we employ the following quantum measurement methods: PauliZ, Z, PauliX, and Probability. \TR{The choice of measurement is critical as it determines the nature of the classical data passed to the subsequent layers of our model.}

\TR{\textit{PauliZ Measurement} -- calculates the expectation value of the Pauli Z operator, which is the average value of its eigenvalues (+1 for state $|0\rangle$ and -1 for state $|1\rangle$). This results in a continuous output value in the range $[-1, +1]$. This continuous output can be interpreted as a probabilistic score or a ``soft'' classification from the quantum circuit. For an SER task, this allows the model to express the degree to which an emotional feature is present, rather than making a hard decision. This nuanced output is then fed into the classical neural network for further processing to reach a final classification~\cite{Zeng2022AStrategy}.} 

\TR{\textit{Z Measurement} -- is a measurement in the computational basis, which forces the qubit into a definite state of either $|0\rangle$ or $|1\rangle$. This process, known as state collapse, yields a classical binary outcome (a 0 or a 1)!\cite{Phalak2024QuaLITi:Performance}. In the context of binary SER classification (e.g., 'High' vs. 'Low' valence), this provides a direct, decisive classification from the quantum layer. This binary output can then be used by the subsequent classical layers, potentially simplifying the final classification decision.}

\TR{The fundamental difference lies in their output: PauliZ provides a continuous, probabilistic value, while Z provides a discrete, binary outcome. This distinction directly affects how the SER model interprets the quantum-processed features, with PauliZ offering a more nuanced, probabilistic representation and Z providing a more direct, binary classification.}

% \textit{PauliZ Measurement} -- corresponds to measuring a qubit in the computational (standard) basis, $\{|0\rangle, |1\rangle\}$. This measurement projects the qubit state onto the eigenstates of the Pauli Z operator: $|0\rangle (eigenvalue +1)$ and $|1\rangle (eigenvalue -1)$. The mathematical representation of the PauliZ measurement is as follows.
% \begin{equation*}
%     Z = \begin{bmatrix} 1 & 0 \\ 0 & -1 \end{bmatrix}
% \end{equation*}

% \textit{Z Measurement} -- is shorthand for a measurement in the Pauli Z basis, projecting the state onto the $|0\rangle$ and $|1\rangle$ eigenstates of the Z operator. It essentially measures
% %BS: added:
% whether 
% %
% the qubit state is $|0\rangle$ or $|1\rangle$, directly giving the classical bit value. The difference between PauliZ and Z measurements are how the results are interpreted or used: as eigenvalues of an operator (+1, -1) or as classical binary outcomes (0, 1).

\textit{PauliX Measurement} -- involves projecting the qubit state onto the eigenstates of the Pauli X operator:
\begin{itemize}
    \item $|+\rangle = \frac{|0\rangle + |1\rangle}{\sqrt{2}} (eigenvalue +1)$
    \item $|-\rangle = \frac{|0\rangle - |1\rangle}{\sqrt{2}} (eigenvalue -1)$
\end{itemize}
This measurement essentially determines the “flip basis” (analogous to flipping the computational basis). The Mathematical representation of the PauliX measurement is as follows.
\begin{equation*}
     X = \begin{bmatrix} 0 & 1 \\ 1 & 0 \end{bmatrix} 
\end{equation*}

\textit{Probability Measurement} -- In quantum mechanics, the probability of obtaining a specific measurement outcome is determined by the Born rule~\cite{Born1926ZurStovorgange}. If a quantum state $|\psi\rangle$ is measured with respect to a specific basis (e.g., Z or X), the probability of observing a particular eigenstate $|\phi\rangle$ is given by Equation~\ref{eq:prob}
\begin{equation}
    \label{eq:prob}
     P(|\phi\rangle) = |\langle \phi | \psi \rangle|^2,
\end{equation}
where $|\langle \phi | \psi \rangle|$ is the overlap between the measured state and the eigenstate.

\subsection{Model Architecture}

Figure~\ref{fig:model_architecture} illustrates the detailed architecture employed in this study. The model builds upon the foundational SER architecture proposed by Issa et al.~\cite{Issa2020SpeechNetworks}. This architecture was selected for its simplicity, as it avoids the complexities introduced by components such as transformers, attention mechanisms, and LSTM networks. By focusing on a streamlined architecture, we aim to isolate and evaluate the specific contribution of the quantum layer to the SER task. 

\TR{We acknowledge the limitations of this approach. The baseline CNN model does not incorporate more complex components like LSTM layers for temporal modeling or attention mechanisms, which have been shown to enhance SER performance by enabling the model to dynamically focus on the most emotionally salient segments of speech~\cite{Chen2021TheRecognition}. This choice represents a critical trade-off: while a more sophisticated classical architecture could potentially yield higher overall accuracy, it would also introduce confounding variables, making it more difficult to isolate and interpret the specific impact of the quantum components. Our primary goal in this study is to establish a clear and foundational understanding of the quantum layer's capabilities. By benchmarking against a simpler, well-understood classical model, we can more directly attribute performance gains to the quantum representation learning, thereby balancing the pursuit of absolute performance with the need for clear, interpretable results.}

The architecture consists of three primary blocks: a CNN-based \textit{classical representation learning block}, a \textit{quantum representation learning block}, and a \textit{classification block}. The input spectrograms are initially processed by the CNN block, which extracts classical feature maps using convolutional operations. These intermediate representations are subsequently forwarded to the quantum representation learning block for further processing. To ensure compatibility between the classical and quantum components, a classical-to-quantum adaptor is introduced. This adaptor reshapes, flattens, and reformats the CNN output to match the input requirements of the quantum embedding layer.

The quantum representation learning block consists of three interconnected modules:
\begin{enumerate}
    \item \textbf{Quantum Embedding Module}: Converts the classical input feature maps into quantum embeddings using quantum embedding algorithms. This step ensures that the input data is represented as quantum states in the Hilbert space, enabling manipulation by quantum gates.
    \item \textbf{Quantum Circuit Layer (\textit{U})}: Utilises \textit{PQCs} to process the quantum embeddings. PQCs include adjustable parameters (e.g., rotation angles) that are optimised during training to enhance the SER task. These circuits enable the model to learn complex and abstract representations by leveraging the unique properties of quantum systems such as superposition and entanglement.
    \item \textbf{Quantum Measurement Module}: Extracts classical information from the processed quantum states by projecting them onto specific basis states. This step produces interpretable outputs, such as probabilities or projections, that are subsequently passed to the classification block.
\end{enumerate}
In this study, 
%BS: you tend to repeat things at times - I added:
as stated, 
we employ four quantum measurement methods -- \textit{PauliZ}, \textit{PauliX}, \textit{Probability}, and \textit{Z} -- to project quantum states onto classical outcomes. These methods are critical for bridging the quantum-to-classical transition, enabling the integration of quantum features into the SER task.

%BS: btw: you arbitrarily reintroduce some abbreviations such as SER while others such as CNN or LSTM are not. Not severe :)
To construct the quantum embedding module, we utilise three embedding algorithms -- \textit{Angle Embedding}, \textit{Amplitude Embedding}, and \textit{IQP Embedding} -- as described in Havl\'{i}\v{c}ek et al.~\cite{Havlicek2019SupervisedSpaces}. These embeddings map input features into quantum vectors of $n$ qubits ($n = 8$ in this study). The resulting quantum vectors are processed through the quantum circuit layer, which applies quantum rotations and other transformations. For this layer, we employ \textit{Strongly Entangling Layers}~\cite{Schuld2020Circuit-centricClassifiers} and \textit{Random Layers} as circuit designs. These circuits facilitate rich entanglement among qubits, enhancing the model's capacity to capture intricate dependencies in the data.

\begin{figure}[h]
    \centering
    \includegraphics[width=1\linewidth]{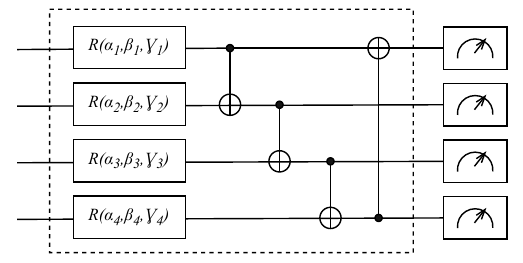}
    \caption{Quantum Circuit Diagram of a Strongly Entangling Layer of four qubits}
    \label{fig:StronglyEntanglingLayer}
\end{figure}

Figure~\ref{fig:StronglyEntanglingLayer} illustrates the quantum circuit of a strongly entangling layer with four qubits. This configuration exemplifies how entanglement is incorporated into the quantum layer to improve the model’s ability to learn nuanced emotional features from speech data.
\TR{The design of this four-qubit circuit, which uses sequential controlled-NOT (CNOT) gates to link the qubits, systematically builds up correlations between them. This robust entanglement is critical for the model's ability to represent the complex, interdependent relationships among the various acoustic features that characterise emotional speech.}

% By integrating quantum representation learning into a classical SER framework, the proposed architecture demonstrates a novel approach for leveraging quantum computing to address the challenges of Speech Emotion Recognition.

%% file: 35_experimental_setup.tex
\section{Experimental Setup}
In this section, we outline the datasets, input features, and model configurations used in this study. The complete Python codebase for the experiments and data pre-processing is publicly accessible on GitHub under the repository \textit{QuantumMachineLearning-for-SER}\footnote{\url{https://github.com/iot-health/QuantumMachineLearning-for-SER}}.

\subsection{Datasets}

% \TR{ write why we are not using MSP-Podcast? }

% We use three publicly available datasets commonly utilised in the field of SER: IEMOCAP~\cite{Busso2008IEMOCAP:Database}, MSP-Improv~\cite{Busso2017MSP-IMPROV:Perception}, and RECOLA~\cite{Ringeval2013IntroducingInteractions}.

\TR{To robustly evaluate the proposed hybrid classical-quantum model, we selected three benchmark datasets commonly used in the SER field: IEMOCAP~\cite{Busso2008IEMOCAP:Database}, MSP-Improv~\cite{Busso2017MSP-IMPROV:Perception}, and RECOLA~\cite{Ringeval2013IntroducingInteractions}}

% \begin{table}[ht]
%     \caption{Summary of the datasets used in this study. The `\# Classes` column contains the total number of classes in the original datasets irrespective of the four emotions (`happiness', `sadness', `anger', and `neutral') we used in our study. \TR{RECOLA Does not have separate Training or Testing } }
%     \label{tab:dataset_summary}
%     \centering
%     \renewcommand{\arraystretch}{1.2}
%     \begin{tabular}{l|l|l|l|l}
% \multicolumn{1}{c|}{\textbf{Dataset}} & \multicolumn{1}{c|}{\textbf{\# Training}} & \multicolumn{1}{c|}{\textbf{\# Testing}} & \multicolumn{1}{c|}{\textbf{\# Classes}} & \multicolumn{1}{c}{\textbf{Language}} \\ \hline
% IEMOCAP    & 2780        & 852        & 4                  & English  \\
% MSP-Improv & 432         & 148        & 4                  & English  \\
% RECOLA        & 000        & 000       & 0                  & English  \\ \hline
% \end{tabular}
% \renewcommand{\arraystretch}{1.0}
%     \end{table}

\subsubsection{IEMOCAP}
\TR{is a widely used dataset containing 12 hours of audio-visual data from dyadic interactions between actors. The sessions include both scripted and improvised scenarios designed to elicit specific emotions. While the emotions are acted, the improvisational nature of many interactions introduces a degree of spontaneity not found in purely scripted datasets. By including IEMOCAP, we test our model's performance on more archetypal and clearly expressed emotions, which serves as a crucial baseline for SER tasks. In this study, we focus on the audio from the improvised sessions and use the valence annotations, which range from 1 to 5, to evaluate the model's ability to capture nuanced emotional states.}
% is a widely used dataset containing 12 hours of acted multi-speaker audio-visual data, with dyadic sessions that include both scripted and improvised scenarios. It provides categorical emotion labels (happiness, sadness, anger, and neutrality) and dimensional labels (valence, dominance, and activation). In this study, we focus on the audio modality from the improvised scenarios and specifically use the valence annotations for analysis. The valance value of this dataset annotation ranges from 1 to 5. 

\subsubsection{MSP-Improv} 
\TR{is a valuable resource designed to bridge the gap between acted and naturalistic data. It features actors improvising scenarios that are specifically designed to elicit genuine emotional responses in a controlled setting. This methodology aims to produce more realistic emotional expressions than simple script-reading, providing a unique test case for our model. The dataset contains 20 predetermined scripts, allowing for the study of key emotions like happiness, sadness, anger, and neutrality. Using MSP-Improv allows us to evaluate our model's effectiveness on lexically controlled but emotionally varied speech.}
% is a widely recognised audio-visual dataset frequently employed in multi modal SER research. Initially developed for a study on audio-visual emotional perception, it has since been utilised in numerous SER studies \cite{Peng2020SpeechFront-Ends, Xie2021SpeechInput, Ahn2021Cross-CorpusAdaptation, Gao2022Domain-InvariantRecognition}, making it a suitable choice for our research. The dataset was recorded in a controlled setting and features 20 pre-determined scripts designed to represent key emotions, including happiness, sadness, anger, and neutrality.

\subsubsection{RECOLA} 
\TR{consists of recordings of spontaneous, naturalistic interactions between participants collaborating on a task remotely. The emotional expressions in RECOLA are not acted or elicited, but arise naturally from the interaction, making them more subtle and representative of real-world conversational speech. This presents a significant challenge for any SER model. By evaluating on RECOLA's valence annotations, we specifically test our model's capacity to recognise emotions in spontaneous and unscripted contexts, which is a key objective for advancing the field of SER.}
% is a multi-modal dataset designed for emotion recognition, containing recordings of naturalistic interactions between pairs of participants. It includes audiovisual and physiological data, annotated with both categorical emotions and dimensional labels, including valence. In this study, we use the valence annotations of RECOLA in our experiments.

\subsection{Input Features}
\label{sec:input_features}
Mel-Frequency Cepstral Coefficients (MFCCs) are commonly used as input features in SER studies~\cite{Likitha2017SpeechMFCC, Latif2019DirectSpeech, Patni2021SpeechFeatures, Dolka2021SpeechFeatures}. While MFCCs capture the spectral characteristics of an audio signal, they do not offer a direct visual representation of the signal's structure. In contrast, Mel-spectrograms provide a two-dimensional visual representation that more intuitively reflects the temporal and spectral patterns of the audio. For this preliminary study, the interpretability of input features in relation to the audio signal is crucial; hence, Mel-spectrograms were chosen as the input features for our model.

To ensure uniformity across the dataset, each audio utterance was resampled to a sampling rate of 22\,kHz and standardised to a duration of 3 seconds, either by truncating or zero-padding as necessary. Mel-spectrograms were then extracted using a window size of 2048 samples and a hop size of 512 samples, resulting in spectrograms with consistent dimensions for input to the model.

\subsection{Model Configuration}
Hyperparameters such as Quantum Measurement methods, Quantum Embedding algorithms, and Quantum Circuit Layers were tuned alongside traditional training parameters, including the optimiser, learning rate, and weight decay. To identify the optimal combination of these hyperparameters, we employed a grid search approach~\cite{Myung2013AOptimization}. A comprehensive summary of the hyperparameters and their respective values used in the grid search process is provided in Table~\ref{tab:grid_search_parameters}.

\begin{table}[ht]
\centering
\caption{Hyper Parameters used in the grid search}
\label{tab:grid_search_parameters}
\renewcommand{\arraystretch}{1.2}
% \resizebox{\linewidth}{!}{
\begin{tabular}{|l|p{5cm}|}
\hline
\multicolumn{1}{|c|}{\textbf{Hyper Parameter}} & \multicolumn{1}{c|}{\textbf{Values}}                \\ \hline
Learning Rate                                             & 0.001, 0.001, 0.00001                               \\ \hline
Optimiser                                      & Adam, SGD, RMSProp, AdaDelta, AdaGrad               \\ \hline
Weight Decay                                             & 0, 0.01, 0.001                                      \\ \hline
Q.Embedding                                    & Angle Embedding, Amplitude Embedding, IQP Embedding \\ \hline
Q.Circuit Layer                                     & Random Layers, Strongly Entangling Layers           \\ \hline
Q.Measurements                                 &  PauliZ, PauliX, Z + PauliZ, Probability             \\ \hline
\end{tabular}
% }
\renewcommand{\arraystretch}{1}
\end{table}

\TR{The models were implemented using the PyTorch framework, with the quantum components simulated using the PennyLane library. All training and evaluation experiments were conducted on a high-performance computing node equipped with an NVIDIA A100 40GB GPU, a 60-core CPU, and 120GB of RAM. Given the extensive hyperparameter space, the grid search was computationally intensive, requiring significant runtime to systematically evaluate all possible configurations and identify the optimal model for each experiment.}

\subsection{Baseline Classical Model}
\begin{figure}[h!]
    \centering
    \includegraphics[width=1\linewidth]{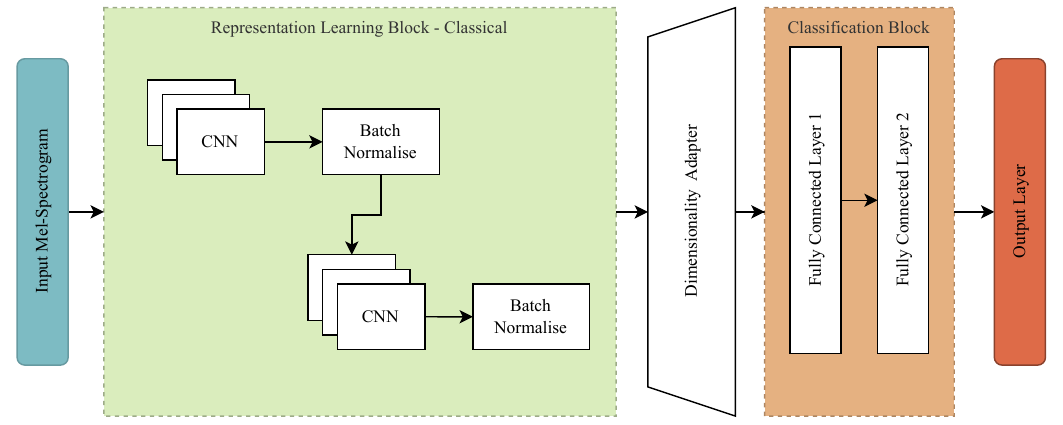}
    \caption{Architecture of the classical model used in this study. The input Mel-spectrogram is processed through a CNN-based representation learning  block, followed by a dimensionality adaptor that feeds into the classification block. The final output layer generates predictions using a softmax function.}
    \label{fig:model_architecture_classical}
\end{figure}

To ensure a fair and rigorous comparison, we performed parallel experiments using a purely classical deep learning approach. The classical DNN model was designed to closely mirror the architecture of the hybrid classical-quantum model, with the key distinction being the exclusion of the `Quantum Representation Learning Block'. The architecture of the classical model used in this study is illustrated in Figure~\ref{fig:model_architecture_classical}.
This experimental setup facilitates a direct comparison between the hybrid classical-quantum approach and traditional deep learning methods, allowing us to isolate and evaluate the impact of the quantum components on model performance.

\subsection{Evaluation Metrics}
We evaluate model performance using the Unweighted Average Recall (UAR, \%), a widely adopted metric in speech-based machine learning research~\cite{Mao2016DomainClasses, Ahn2021Cross-CorpusAdaptation, Ishaq2023TC-Net:Network, Khan2024MSER:Fusion}. UAR is calculated by first computing the recall for each class label in the classification task and then averaging these recall values equally across all labels.

\subsection{Experiments}
We designed our experiments to address two primary scenarios: binary classification and multi-class emotion classification.

\begin{itemize} \item \textbf{Binary Classification:} The objective of this task was to classify input speech samples into one of two categories: ``High'' or ``Low'' valence. For this evaluation, we utilised the IEMOCAP and RECOLA datasets, which are well-suited for valence-based classification. \item \textbf{Multi-Class Emotion Classification:} This task aimed to identify the specific emotion conveyed in the input speech, selecting from the categories: happy, angry, sad, or neutral. For this purpose, we employed the IEMOCAP and MSP-Improv datasets, which offer comprehensive annotations for a diverse range of emotions. \end{itemize}

%% file: 40_results.tex
\section{Evaluation}
\label{sec:results}

% In this section we discuss the experiments we did during our study and the results of those experiments. We mainly did experiments under two scenarios, Binary classification and multi-class emotion classification. Results of the binary classification are mentioned in the subsection~\ref{sec:binary_classification} while the results of the multi-class classification is mentioned in the subsection~\ref{sec:4class_classifiaction}.

% For the apple-to-apple comparison, we performed an identical experiment in classical scenario in this study. The DNN model of the classical scenario is exactly as the classical-quantum model, with the exception of "Quantum Feature Extraction Block". 

% This way we can compare the Hybrid classical-quantum approach with classical deep learning approaches. 

% We show the UAR\% as a factor of performance while we also show the number of trainable parameters in each model as an indication of how complex the models are. 

% This section presents the experiments conducted in our study and their corresponding results. The experiments performed under the two scenarios: binary classification and multi-class emotion classification. The results of the binary classification are detailed in subsection~\ref{sec:binary_classification}, while those of the multi-class classification are provided in subsection~\ref{sec:4class_classifiaction}.

This section presents the results of the experiments conducted in our study. The results of the binary classification are detailed in subsection~\ref{sec:binary_classification}, while those of the multi-class classification are provided in subsection~\ref{sec:4class_classifiaction}.
Performance is evaluated using UAR\%, while the number of trainable parameters is reported to indicate the relative complexity of each model.

\begin{figure*}[t]
    \centering
    \includegraphics[width=.8\textwidth]{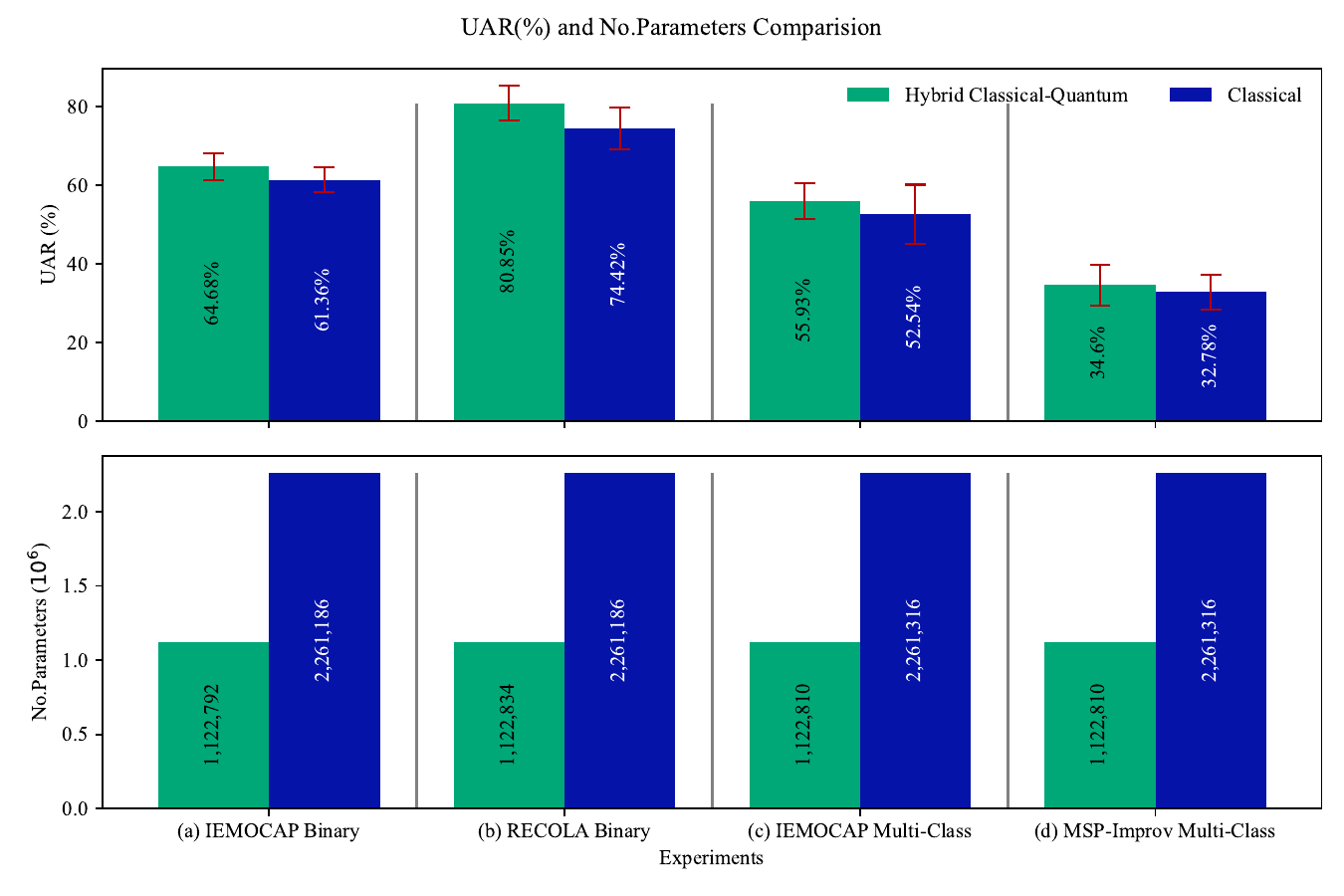}
    \caption{Comparison of UAR (\%) and Number of Parameters with Hybrid classical-quantum model and Classical Model in the experiments (a) IEMOCAP Binary Classification, (b) RECOLA Binary Classification, (c) IEMOCAP Multi-class Classification, and (d) MSP-Improv Multi-class Classification. }
    \label{fig:results_combined}
\end{figure*}

\input{40_results_summary}

\subsection{Binary Classification}
\label{sec:binary_classification}

In this section, we evaluate the binary classification performance of the hybrid classical-quantum models within the SER domain. The experiments were conducted using the IEMOCAP and RECOLA datasets. The objective was to classify speech emotion into one of two dimensional labels: Valence -- High or Low.

\subsubsection{IEMOCAP}
The IEMOCAP dataset is annotated with both categorical and dimensional emotional labels. For the valence dimension, annotations range from 1 to 5. We categorised valence as “Low” for values less than 3 and “High” for values 3 and above.

% \begin{figure}[ht]
%     \centering
%     \includegraphics[width=1\linewidth]{figures/IEM_2Class.pdf}
%     \caption{UAR\% and Number of parameters of the Best Quantum Model and it's corresponding Classical Model for IEMOCAP dataset for binary classification.}
%     \label{fig:IEM_2Class}
% \end{figure}
Figure~\ref{fig:results_combined}(a) illustrates the UAR\% and the number of trainable parameters for the best-performing model obtained through grid search. 
The hyper-parameters selected from the grid search for the best performing model are:  
\begin{itemize}
    \item Learning Rate: 0.00001 
    \item Optimiser: Adam Optimiser with zero weight decay
    \item Quantum Layer: Angle Embedding, Random Layers as Quantum Circuit and Summation of Z and PauliZ measurements as Quantum Measurement. 
\end{itemize}
It also shows the comparison with the identical classical model. 

Analysing the UAR\% results, the hybrid classical-quantum model outperforms the identical classical DNN model. This suggests that the inclusion of the ``Quantum representation learning block'' contributes significantly to the observed performance improvement~\cite{Thejha2023SpeechNetwork,Esposito2022QuantumHealthcare}.
% \TR{add rationale from the literature }

Also, the hybrid classical-quantum model has nearly as half of the number of parameters which indicates the reduced training complexity of the model. 

\subsubsection{RECOLA}
The RECOLA dataset is an annotated resource containing multiple markers for emotion recognition. For this binary classification task in SER, we focused on the valence dimension. Valence in the RECOLA dataset is annotated with positive and negative values. We categorised negative valence values as ``Low'' and positive valence values as ``High''.

% \begin{figure}[ht]
% \centering
% \includegraphics[width=1\linewidth]{figures/RECOLA_2Class.pdf}
% \caption{UAR\% and number of parameters for the best-performing quantum model and its corresponding classical model on the RECOLA dataset for binary classification.}
% \label{fig:RECOLA_2Class}
% \end{figure}

Figure~\ref{fig:results_combined}(b) presents the UAR\% and the number of trainable parameters for each hybrid classical-quantum model and its corresponding classical model. The best-performing quantum model, selected via grid search, is configured with the following hyper-parameters:
\begin{itemize}
\item Learning Rate: 0.00001
\item Optimiser: Stochastic Gradient Descent (SGD) with zero weight decay
\item Quantum Layer: Amplitude Embedding, Strongly Entangling Layers for the Quantum Circuit, and PauliX for Quantum Measurement
\end{itemize}

From the results, the hybrid classical-quantum model demonstrates superior performance compared to its classical counterpart, achieving this with nearly half the number of trainable parameters.

\subsection{Multi-class Emotion Classification}
\label{sec:4class_classifiaction}

In this section, we assess the multi-class emotion classification performance of the hybrid classical-quantum models in the SER domain. The experiments were conducted using the IEMOCAP and MSP-Improv datasets. The goal was to classify speech emotions into one of four categorical labels: Angry, Happy, Neutral, and Sad.

% \subsubsection{IEMOCAP}
% We use the categorical annotation of the IEMOCAP dataset in this scenario. 

% \begin{figure}[ht]
%     \centering
%     \includegraphics[width=1\linewidth]{figures/IEM_4Class.pdf}
%     \caption{UAR\% and Number of parameters of the Best Quantum Model and it's corresponding Classical Model for IEMOCAP dataset for multi-class classification.}
%     \label{fig:IEM_4Class}
% \end{figure}
% The Figure~\ref{fig:IEM_4Class} show the results of the best quantum model vs the corresponding classical model. 

% Comparing the Figure~\ref{fig:IEM_2Class} and Figure~\ref{fig:IEM_4Class}, it is seen that the UAR\% of the multi-class classification task has lesser performance. This is due to the complexity of the SER task and the increment of number of output classes. On the other-hand, the quantum model in this scenario has higher accuracy compared to the corresponding classical model. 

% The selected model has the hyper-parameters; 
% \begin{itemize}
%     \item Learning Rate: 0.001
%     \item Optimiser: Stochastic Gradient Descent (SGD) with zero weight decay
%     \item Quantum Layer: Quantum Layer: Angle Embedding, Random Layers as Quantum Circuit and Summation of Z and Pauli Z measurements as Quantum Measurement. 
% \end{itemize}
% Comparing with the Binary Classification experiments with IEMOCAP, it is visible that the quantum layer selected for both the scenarios (Binary \& Multi-class) are the same. This indicates the parameters of the quantum layer should be selected based on the data distribution. 

\subsubsection{IEMOCAP}
In this scenario, we utilise the categorical annotations of the IEMOCAP dataset.

% \begin{figure}[ht]
% \centering
% \includegraphics[width=1\linewidth]{figures/IEM_4Class.pdf}
% \caption{UAR\% and number of parameters for the best quantum model and its corresponding classical model on the IEMOCAP dataset for multi-class classification.}
% \label{fig:IEM_4Class}
% \end{figure}

Figure~\ref{fig:results_combined}(c) presents the performance comparison between the best quantum model and its classical counterpart.

When comparing Figure~\ref{fig:results_combined}(a) and Figure~\ref{fig:results_combined}(c), it is evident that the UAR\% for the multi-class classification task is lower. This decline in performance can be attributed to the increased complexity of the SER task and the higher number of output classes. However, the quantum model demonstrates higher accuracy compared to the corresponding classical model in this multi-class scenario.

The selected model employs the following hyper-parameters:
\begin{itemize}
\item Learning Rate: 0.001
\item Optimiser: Stochastic Gradient Descent (SGD) with zero weight decay
\item Quantum Layer: Angle embedding, random layers as the quantum circuit, and a combination of Z and PauliZ measurements for quantum measurement
\end{itemize}

When compared to the binary classification experiments on the IEMOCAP dataset, the same quantum layer configuration is observed for both binary and multi-class scenarios. This suggests that the parameters of the quantum layer should be tailored to the underlying data distribution.

\subsubsection{MSP-Improv}
We use four categorical emotions (Angry, Happy, Neutral, and Sad) annotated in the MSP-Improve dataset. The task of this classification model is to correctly classify the emotion embedded in the audio. 

% \begin{figure}[ht]
%     \centering
%     \includegraphics[width=1\linewidth]{figures/MSP_IMPROV_4Class.pdf}
%     \caption{UAR\% and number of parameters for the best quantum model and its corresponding classical model on the MSP-Improve dataset for multi-class classification.}
%     \label{fig:MSP_IMPROV_4Class}
% \end{figure}
Figure~\ref{fig:results_combined}(d) compares the performance and the complexity of the best quantum model selected by the grid search and the corresponding classical model. The hyper-parameters selected by grid search for the best performing quantum model are: 
\begin{itemize}
    \item Learning Rate: 0.0001
    \item Optimiser: AdaGrad optimiser with zero weight decay
    \item Quantum Layer: Angle embedding, random layers as the quantum circuit, and PauliZ measurements for quantum measurement
\end{itemize}

\subsection{Summary of the Results}
A summary of the results obtained by the experiments in this study is tabulated in the Table~\ref{tab:results_summary}.
The results demonstrate the effectiveness of the hybrid classical-quantum model for SER tasks, highlighting superior performance and reduced complexity compared to classical models. For binary classification on the IEMOCAP dataset, the hybrid model achieved a UAR of 64.68\%, outperforming the classical model's 61.36\%, with 50.34\% fewer trainable parameters. 

Similarly, for the RECOLA dataset, the hybrid model recorded a UAR of 80.85\%, %BS: do not write "significantly" unless naming a test method and p-value - I changed:
considerably 
%
%BS: You report XX.XX% results - this requires at least 10,000 test instances which you did not have - professionally, it should be reported as XX.X% - let's leave now, however, as is :)
higher than the classical model's 74.42\%, while maintaining the same parameter efficiency. In multi-class classification, the hybrid model also outperformed its classical counterpart, achieving a UAR of 55.93\% on the IEMOCAP dataset (versus 52.54\%) and 34.60\% on MSP-Improv (versus 32.78\%), again with reduced complexity. Across all tasks, optimal configurations for the hybrid model consistently included zero weight decay and specific quantum embeddings, such as angle or amplitude embedding, paired with random or strongly entangling quantum layers. These results highlight the potential of integrating quantum feature extraction into SER models, offering improved accuracy and efficiency in handling complex emotional data.

%% file: 40_results_summary.tex
\begin{table*}[t]
\caption{UAR (\%) of Best Quantum Model and corresponding Classical Model along with the Hyper-parameters selected by Grid Search.}
\label{tab:results_summary}
\centering
\renewcommand{\arraystretch}{1.2}
\begin{tabular}{l|llll|}
\cline{2-5}
\multirow{2}{*}{\textbf{Hyper Parameters}} & \multicolumn{4}{c|}{\textbf{Dataset}}                                                                                                                                              \\
                                           & \multicolumn{1}{l|}{\textbf{IEMOCAP - Binary}} & \multicolumn{1}{l|}{\textbf{RECOLA - Binary}}   & \multicolumn{1}{l|}{\textbf{IEMOCAP - 4 Class}} & \textbf{MSP-Improv - 4 Class} \\ \hline
Learning Rate                              & \multicolumn{1}{l|}{0.00001}                   & \multicolumn{1}{l|}{0.00001}                    & \multicolumn{1}{l|}{0.001}                      & 0.0001                        \\
Optimiser                                  & \multicolumn{1}{l|}{Adam}                      & \multicolumn{1}{l|}{SGD}                        & \multicolumn{1}{l|}{SGD}                        & AdaGrad                       \\
Weight Decay                               & \multicolumn{1}{l|}{0}                         & \multicolumn{1}{l|}{0}                          & \multicolumn{1}{l|}{0}                          & 0                             \\
Q.Embedding                                & \multicolumn{1}{l|}{Angle Embedding}           & \multicolumn{1}{l|}{Amplitude Embedding}        & \multicolumn{1}{l|}{Angle Embedding}            & Angle Embedding               \\
Q.Circuit Layer                                 & \multicolumn{1}{l|}{Random Layers}             & \multicolumn{1}{l|}{Strongly Entangling Layers} & \multicolumn{1}{l|}{Random Layers}              & Random Layers                 \\ 
Q.Measurement                              & \multicolumn{1}{l|}{Z + PauliZ}                & \multicolumn{1}{l|}{PauliX}                     & \multicolumn{1}{l|}{Z + PauliZ}                 & PauliZ                        \\ \hline
UAR (\%) - Quantum                         & \multicolumn{1}{l|}{\textbf{64.68 $\pm$ 3.34}}          & \multicolumn{1}{l|}{\textbf{80.85 $\pm$ 4.45}}           & \multicolumn{1}{l|}{\textbf{55.93 $\pm$ 4.62}}           & \textbf{34.60 $\pm$ 5.19}              \\
UAR (\%) - Classical                       & \multicolumn{1}{l|}{61.36 $\pm$ 3.21}          & \multicolumn{1}{l|}{74.42 $\pm$ 5.28}           & \multicolumn{1}{l|}{52.54 $\pm$ 7.56}           & 32.78 $\pm$ 4.32              \\ \hline
\end{tabular}
\renewcommand{\arraystretch}{1.0}
\end{table*}

%% file: 50_discussion.tex
\section{Discussion and Future Work}

% The results of the study highlight how well the proposed hybrid classical-quantum framework can handle the challenges that come with SER. The work shows that combining PQCs with a traditional CNN substantially decreases model complexity while improving classification performance on both binary and multi-class problems.

% For binary classification tasks, the hybrid models demonstrated higher accuracies on both the IEMOCAP and RECOLA datasets, consistently outperforming their classical counterparts in terms of UAR(\%). The quantum layer's potential to utilise entanglement and superposition, which enhance feature representation and capture subtle relationships among speech components, contributes for these improvements. Furthermore, the hybrid approach's effectiveness is demonstrated by a reduction in the number of trainable parameters, which makes it an intriguing option for situations with limited resources.

% The hybrid framework had slightly lower UAR(\%) values than binary tasks, but it still performed better for multi-class classification. The higher complexity involved in identifying more subtle emotional states is consistent with this outcome. The quantum-enhanced model proved stable in spite of this difficulty, especially when tested on the MSP-Improv dataset, which includes a variety of emotional expressions.

The results of the study highlight how well the proposed hybrid classical-quantum framework can handle the challenges that come with SER. The work shows that combining PQCs with a traditional CNN substantially decreases model complexity while improving classification performance on both binary and multi-class problems.

For binary classification tasks, the hybrid models demonstrated higher accuracies on both the IEMOCAP and RECOLA datasets, consistently outperforming their classical counterparts in terms of UAR(\%). The quantum layer's potential to utilise entanglement and superposition, which enhance feature representation and capture subtle relationships among speech components, contributes for these improvements. Furthermore, the hybrid approach's effectiveness is demonstrated by a reduction in the number of trainable parameters, which makes it an intriguing option for situations with limited resources.

The hybrid framework had slightly lower UAR(\%) values than binary tasks, but it still performed better for multi-class classification. The higher complexity involved in identifying more subtle emotional states is consistent with this outcome. The quantum-enhanced model proved stable in spite of this difficulty, especially when tested on the MSP-Improv dataset, which includes a variety of emotional expressions.

\subsection{Key Observations}
\label{sec:key_observations}
Key observations observed throughout this study are;
\begin{itemize}
    \item The experiments reveal that quantum embeddings and circuit configurations (e.g., Angle Embedding and Strongly Entangling Layers) are critical to achieving optimal performance. The observed consistency in selected quantum components across binary and multi-class tasks suggests a potential universality of these configurations for SER tasks.
    \item The absence of weight decay in the best-performing models indicates that quantum layers inherently provide sufficient regularisation.
    \item The hybrid models' reduced parameter count underscores their suitability for deployment in real-world scenarios where computational resources and energy efficiency are critical constraints.
\end{itemize}

\subsection{Lessons Learnt}
The development of the hybrid classical-quantum framework presented in this paper involved a significant exploration of different architectural and methodological approaches. Before arriving at the final design, several alternative strategies were implemented and evaluated, providing valuable insights into the challenges and opportunities of integrating QML with SER. These explorations highlighted the complexities of leveraging quantum properties effectively and guided the iterative refinement of the model towards improved performance and efficiency.

\begin{figure}[t]
    \begin{subfigure}{.5\linewidth}
        \centering
        \includegraphics[width=1\linewidth]{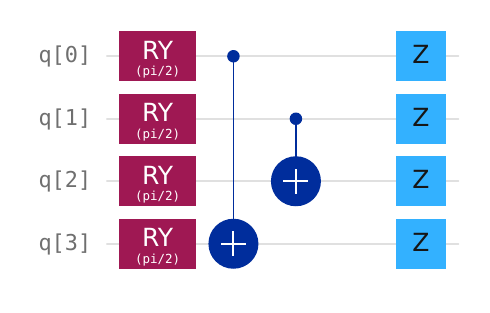}
        \caption{2-CNOT}
        \label{fig:sub1}
    \end{subfigure}%
    \begin{subfigure}{.5\linewidth}
        \centering
        \includegraphics[width=1\linewidth]{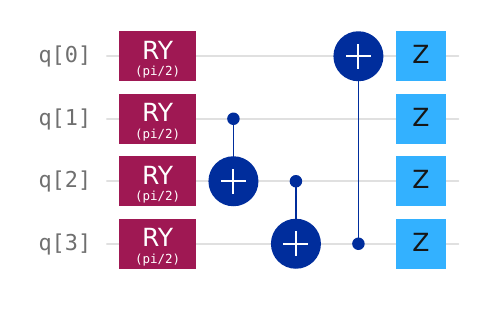}
        \caption{3-CNOT}
        \label{fig:sub2}
    \end{subfigure}\\[1ex]
    \begin{subfigure}{1\linewidth}
        \centering
        \includegraphics[width=.5\linewidth]{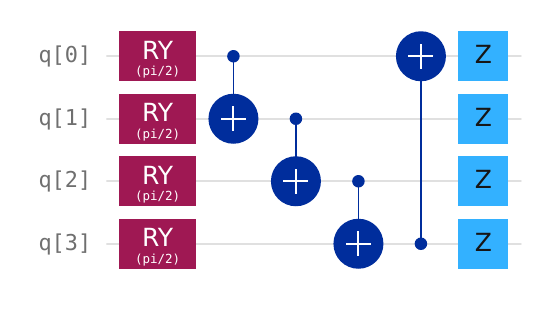}
        \caption{4-CNOT}
        \label{fig:sub3}
    \end{subfigure}
    \caption{2-CNOT, 3-CNOT, and 4-CNOT quantum circuits we used in the experiments}
    \label{fig:static_q_circuits}
\end{figure}

\begin{figure*}[t]
    \centering
    \begin{subfigure}{0.85\textwidth}
        \centering
        \includegraphics[width=1\linewidth]{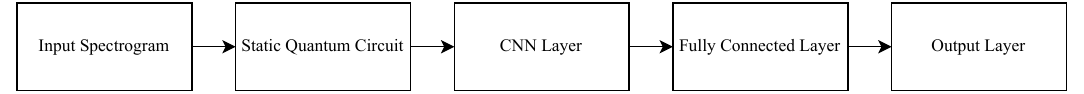}
        \caption{CNN-based SER model architecture employing a static quantum circuit as the feature extractor}
        \label{fig:sub_CNN_SER}
    \end{subfigure}%
    \\[2ex]
    \begin{subfigure}{0.85\textwidth}
        \centering
        \includegraphics[width=1\linewidth]{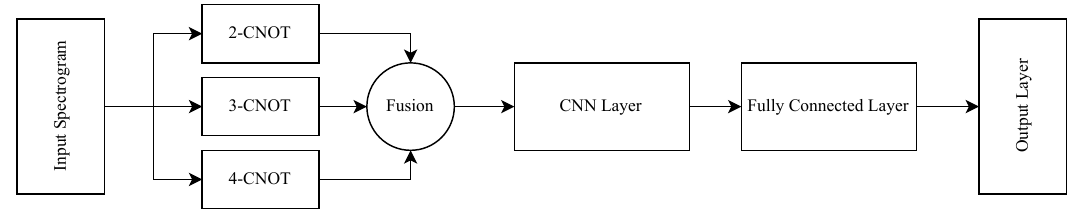}
        \caption{A fusion-based SER model architecture employing three static quantum circuits to extract and early-fuse the features.}
        \label{fig:sub_ealry_fusion}
    \end{subfigure}
    \caption{Two of the model architectures employed in our experiments involving static quantum circuits.}
    \label{fig:static_q_circuit_models}
\end{figure*}

Our initial investigations focused on incorporating static quantum circuits within established deep learning architectures. We hypothesised that these circuits, despite their fixed structure, could offer advantages in feature extraction due to their ability to exploit quantum phenomena like superposition. A variety of circuit configurations, including those based on 2-CNOT, 3-CNOT, and 4-CNOT gates (Figure~\ref{fig:static_q_circuits}), were designed and tested. The output of these static quantum circuits, essentially quantum measurements representing extracted features, were then passed as input to various classical deep learning models. These included standard CNNs, LSTMs for capturing temporal dependencies, multi-kernel CNNs to explore different receptive fields, and combinations of CNNs and LSTMs (Figure~\ref{fig:static_q_circuit_models} (a)). However, across these various architectures and circuit designs, the performance on benchmark datasets like IEMOCAP remained consistently below 49\% accuracy. This suggested that the static nature of these circuits limited their ability to adapt to the nuances and complexities of emotional expression in speech data.

Subsequently, we explored the potential of fusion-based models, aiming to combine the strengths of both quantum and classical processing. Two primary fusion strategies were investigated: early fusion and decision-level fusion. In the early fusion approach, the outputs of the static quantum circuits and the classical CNNs were concatenated before being passed to a final classification layer (Figure~\ref{fig:static_q_circuit_models} (b)). This aimed to leverage both quantum and classical features in the decision-making process. The decision-level fusion approach, on the other hand, involved training separate quantum and classical models, and their respective classification outputs were then combined using techniques like averaging or weighted averaging. Despite the conceptual appeal of these fusion strategies, the observed performance gains were marginal, with accuracies hovering around 44\% on the IEMOCAP dataset. This suggested that simply combining static quantum computations with classical deep learning was insufficient to capture the intricate relationships within emotional speech data.

Further experimentation delved into the role of regularisation within these hybrid models. Specifically, we investigated the impact of L2 regularisation, commonly used to prevent over-fitting in classical deep learning. We observed a clear trend: increasing the weight decay (the hyperparameter controlling L2 regularisation) improved the performance of the classical CNN components. However, the same trend was not observed for the quantum components, suggesting that they might possess inherent regularisation properties due to the constraints of the quantum Hilbert space.

The culmination of these explorations led to the realisation that the fixed nature of static quantum circuits was a limiting factor in achieving optimal performance. This prompted the shift towards PQCs, which allow for adaptive learning through the optimisation of circuit parameters. The flexibility of PQCs, combined with carefully chosen quantum embeddings and circuit architectures, proved to be the key to unlocking the potential of QML for SER, as demonstrated by the improved results presented in this study. This journey emphasises the significance of systematic exploration and iterative refinement in the nascent field of QML, laying the groundwork for future research to further optimise and expand upon these promising initial discoveries.

% \TR{To come here we have done couple of things like Static Q.Circuits, multi head NN}

% \TR{Talk about the experiments we did (what we tried and and their results). e.g (static Q. Circuits)  }

\subsection{Limitations of the Proposed Model}
While the hybrid framework offers promising results, several limitations merit discussion. First, the study relies on simulated quantum environments, which may not fully capture the hardware-related challenges and noise associated with physical quantum systems. Future work should validate these findings on real quantum devices to assess their practical applicability.

Additionally, the current approach uses fixed configurations for the classical CNN component. Exploring alternative architectures, such as transformer-based models, could further enhance performance by leveraging their strengths in capturing long-term dependencies.

Finally, the scope of datasets used in this study, though diverse, does not encompass all linguistic and cultural variations in emotional speech. Expanding the evaluation to include more diverse datasets would provide a more comprehensive understanding of the model's generalisation capacity.

%% file: 60_conclusion.tex
\section{Conclusion}

This paper has introduced a novel hybrid classical-quantum framework for Speech Emotion Recognition (SER), leveraging the power of Parameterised Quantum Circuits (PQCs) integrated within a Convolutional Neural Network (CNN) architecture. By harnessing the principles of quantum superposition and entanglement, the proposed model demonstrates a significant improvement in both accuracy and model efficiency. 

The experimental results across three widely used benchmark datasets -- IEMOCAP, RECOLA, and MSP-Improv -- consistently showcase the superior performance of the hybrid classical-quantum model compared to its purely classical counterpart. Notably, the hybrid models achieve enhanced accuracy in both binary and multi-class emotion classification tasks, while simultaneously reducing the number of trainable parameters by 50.34\%. This reduction in model complexity translates to lower computational overhead and improved energy efficiency, rendering these quantum-enhanced methods more suitable for practical real-world applications.

Furthermore, the study reveals crucial insights into optimal configurations for hybrid classical-quantum models in SER. The experiments suggest that specific quantum embeddings, such as Angle and Amplitude embedding, coupled with strongly entangling or random quantum circuit layers, and a summation of Pauli Z measurements consistently contribute to superior performance. 
The absence of weight decay in these optimal configurations implies that L2 regularisation might not be necessary for such hybrid models. 
\TR{Notably, the optimal configurations consistently emerge with zero weight decay. This directly supports the observation from Section~\ref{sec:key_observations}, which indicates that the quantum layers inherently provide sufficient regularisation. This behaviour suggests that classical techniques like L2 regularisation may be redundant, potentially due to the intrinsic properties and constraints of the quantum Hilbert space, simplifying the training process.}
These findings provide valuable guidance for future research and development in quantum-enhanced SER.

While these results are promising, the study also acknowledges certain limitations. Firstly, the experiments were performed using simulated quantum environments. Future studies should validate these findings on physical quantum devices. Secondly, the current implementation uses a fixed architecture for the classical CNN component; exploring alternative architectures could be an avenue for further improvement. Finally, the dataset diversity should be expanded in future studies to better understand the models' generalisation capacity across various linguistic and cultural contexts.

In summary, this study demonstrates the potential of quantum machine learning to revolutionise the field of SER. By integrating PQCs with classical CNNs, our proposed framework achieves improved accuracy and reduced complexity, paving the way for the development of more robust and efficient SER models for real-world applications. The future work includes extending these experiments on a quantum hardware and implementing more robust feature extraction methods for the CNN model.